\documentclass[sigconf]{acmart}
\AtBeginDocument{%
  }

\copyrightyear{2026}
\acmYear{2026}
\setcopyright{cc}
\setcctype{by}
\acmConference[CIKM '26]{Proceedings of the 35th ACM International Conference on Information and Knowledge Management}{November 07--11, 2026}{Rome, Italy}
\acmBooktitle{Proceedings of the 35th ACM International Conference on Information and Knowledge Management (CIKM '26), November 07--11, 2026, Rome, Italy}
\acmDOI{10.1145/3799682.3840072}
\acmISBN{979-8-4007-2539-5/2026/11}

\usepackage{multirow}

\begin{document}

%% The "title" command has an optional parameter,
%% allowing the author to define a "short title" to be used in page headers.
\title{Beyond One-Shot Expansion: Contrastive Evidence Exploration for Multi-Hop Retrieval}

%% The "author" command and its associated commands are used to define
%% the authors and their affiliations.
%% Of note is the shared affiliation of the first two authors, and the
%% "authornote" and "authornotemark" commands
%% used to denote shared contribution to the research.
\author{JungMin Yun}
\orcid{0000-0001-6868-286X}
\affiliation{%
  \institution{Chung-Ang University}
  \city{Seoul}
  \country{Republic of Korea}
}
\email{cocoro357@cau.ac.kr}

\author{YoungBin Kim}
\orcid{0000-0002-2114-0120}
\authornote{Corresponding author}
\affiliation{%
  \institution{Chung-Ang University}
  \city{Seoul}
  \country{Republic of Korea}
}
\email{ybkim85@cau.ac.kr}

%% By default, the full list of authors will be used in the page
%% headers. Often, this list is too long, and will overlap
%% other information printed in the page headers. This command allows
%% the author to define a more concise list
%% of authors' names for this purpose.
%\renewcommand{\shortauthors}{Trovato et al.}
\renewcommand{\shortauthors}{JungMin Yun and YoungBin Kim}

%% The abstract is a short summary of the work to be presented in the article.
\begin{abstract}
  Retrieval-augmented generation (RAG) critically depends on retrieving the evidence necessary for effective reasoning. However, this remains particularly challenging in multi-hop question answering (QA), where supporting passages are often linked through intermediate entities and relations that must be progressively uncovered. Existing retrieval approaches typically rely on a single retrieval intent or one-shot query expansion, limiting their ability to adapt to newly retrieved evidence and potentially introducing noisy or redundant retrieval signals. To address these limitations, we propose a training-free multi-hop retrieval framework that integrates evidence-conditioned exploration, passage-specific contrastive refinement, and coverage-aware final ranking. During offline indexing, the framework constructs passage-specific contrastive facets that characterize each passage relative to its semantically similar neighbors, providing fine-grained signals to distinguish closely related candidates. At inference time, the framework iteratively retrieves evidence, generates probes targeting unresolved information needs, refines candidate relevance using the contrastive facets, and selects a complementary set of passages that collectively cover diverse evidence-seeking intents. Experiments on MuSiQue, HotpotQA, and 2WikiMultihopQA demonstrate consistent improvements in retrieval quality and downstream QA performance over baselines.
\end{abstract}



%% Keywords. The author(s) should pick words that accurately describe
%% the work being presented. Separate the keywords with commas.

\maketitle

\section{Introduction}

Retrieval-augmented generation (RAG) has become a standard paradigm for knowledge-intensive question answering (QA)~\cite{lewis2020retrieval, gao2023retrieval, zhao-etal-2026-r-3ag, wang-etal-2025-rag}, yet its effectiveness critically depends on retrieving the evidence required for reasoning~\cite{sawarkar2024blended, kambattan2025ai, zhuang2024efficientrag}. This is particularly challenging in multi-hop QA, where answering a question often requires connecting multiple passages through intermediate entities, relations, or reasoning steps that are not explicitly stated in the original question~\cite{trivedi-etal-2022-musique, zhu2025mitigating, krishna-etal-2025-fact}. However, many existing retrieval methods still treat the input question as a single retrieval intent. Consequently, they may prioritize passages that are directly similar to the question while overlooking complementary evidence needed to complete the reasoning chain~\cite{yun2025query, zhuang2024efficientrag, petcu-etal-2026-query}.

Large language model (LLM)-based query expansion addresses this limitation by enriching the original question with generated documents, reformulated queries, or auxiliary retrieval cues~\cite{gao-etal-2023-precise, wang-etal-2023-query2doc, zhuang2024efficientrag, shen-etal-2024-retrieval}. Although such methods can provide useful contextual signals and improve retrieval recall, they are often applied as a static, one-shot expansion before retrieval~\cite{gao-etal-2023-precise, wang-etal-2023-query2doc, kim-etal-2026-qudar}. As a result, they have limited capacity to adapt the retrieval direction to evidence uncovered during the retrieval process~\cite{trivedi2023interleaving, shao2023enhancing, li-etal-2026-retrieval}. In multi-hop QA, however, the information needed at a subsequent step often depends on the evidence retrieved at earlier steps~\cite{wang2025chainofretrieval, wei-etal-2026-cirag}. This dependency calls for an evidence-aware and adaptive exploration process that can dynamically capture emerging information needs as retrieval progresses.

Iterative retrieval methods move toward this goal by interleaving retrieval and reasoning and using generated reasoning steps, follow-up questions, or structured knowledge connections to guide subsequent retrieval~\cite{trivedi2023interleaving, shao2023enhancing, zhuang2024efficientrag, press-etal-2023-measuring}. Nevertheless, these intermediate signals may remain noisy, incomplete, or insufficiently grounded in the retrieved evidence~\cite{chu2024beamaggr, zhu2025mitigating, li-etal-2026-s2g, amiraz2025distracting}, potentially leading subsequent retrieval toward irrelevant or already resolved information needs. Moreover, even when relevant evidence is successfully retrieved, the final retrieved set may still contain substantial redundancy if ranking considers passages primarily in isolation and does not explicitly account for the coverage of distinct evidence-seeking intents~\cite{min2021joint, lee-etal-2025-shifting, in-etal-2025-diversify, sun-etal-2025-enhancing-retrieval}.

To address these challenges, we argue that effective multi-hop retrieval should jointly consider three complementary aspects: \textit{adaptive exploration}, \textit{passage-specific refinement}, and \textit{evidence coverage}. Rather than relying solely on the original question or a one-shot expansion, the retriever should progressively explore new evidence directions as retrieval proceeds, allowing newly discovered entities and relations to guide subsequent search. It should also distinguish among semantically similar passages that support different entities or relations, thereby preserving fine-grained evidence distinctions. The retriever should select passages that collectively cover the diverse evidence-seeking intents required to complete the reasoning chain, rather than repeatedly favoring evidence that serves the same retrieval purpose.

Motivated by this perspective, we propose a training-free multi-hop retrieval framework that progressively explores evidence conditioned on the retrieved context and selects complementary supporting passages. During offline indexing, the framework constructs passage-specific contrastive facets in the form of queries that distinguish each passage from its semantically similar neighbors. At inference time, it sequentially retrieves evidence and generates probes targeting unresolved information needs. The accumulated candidates are then refined using the contrastive facets and ranked with a coverage-aware objective that favors passages providing complementary evidence.

Our contributions are threefold. First, we formulate multi-hop retrieval as an adaptive evidence-exploration process in which retrieval directions evolve with the evidence discovered so far, rather than remaining fixed by the original question or a one-shot expansion. Second, we propose a training-free multi-hop retrieval framework that integrates contrastive facet indexing, evidence-conditioned exploration, passage-specific relevance refinement, and coverage-aware selection. Third, experiments on MuSiQue, HotpotQA, and 2WikiMultihopQA demonstrate consistent improvements in retrieval quality and downstream QA performance.

\section{Methodology}

\subsection{Problem Formulation}

Let $q$ denote an input multi-hop question, and let $\mathcal{D}=\{d_1,\ldots,d_N\}$ denote a corpus of passages. The goal of multi-hop retrieval is to return a ranked set of $K$ passages that collectively provide the evidence required to answer $q$. Unlike single-hop retrieval, relevant passages in multi-hop settings may correspond to intermediate entities, relations, or reasoning steps that are not explicitly expressed in the surface form of the original question. An effective retriever should therefore identify not only passages that are individually relevant to $q$, but also complementary evidence that supports different stages of the reasoning chain.

We assume access to a pretrained text encoder that maps queries and passages into a dense representation space. Let $\phi(x)$ denote the $\ell_2$-normalized embedding of text $x$. The semantic relevance between two texts $x$ and $y$ is measured by cosine similarity, defined as $\operatorname{sim}(x,y)=\phi(x)^\top\phi(y)$. A standard dense retriever ranks passages according to $\operatorname{sim}(q,d)$.

However, relying on a single representation of the original question can be insufficient in multi-hop retrieval, particularly when supporting passages contain latent bridge entities or intermediate relations that are only revealed after partial evidence has been retrieved. To address this limitation, we propose a training-free multi-hop retrieval framework that integrates contrastive facet indexing, sequential evidence exploration, contrastive evidence refinement, and coverage-aware final ranking.

\begin{figure}[t!]
  \centering
\includegraphics[width=1.0\columnwidth]{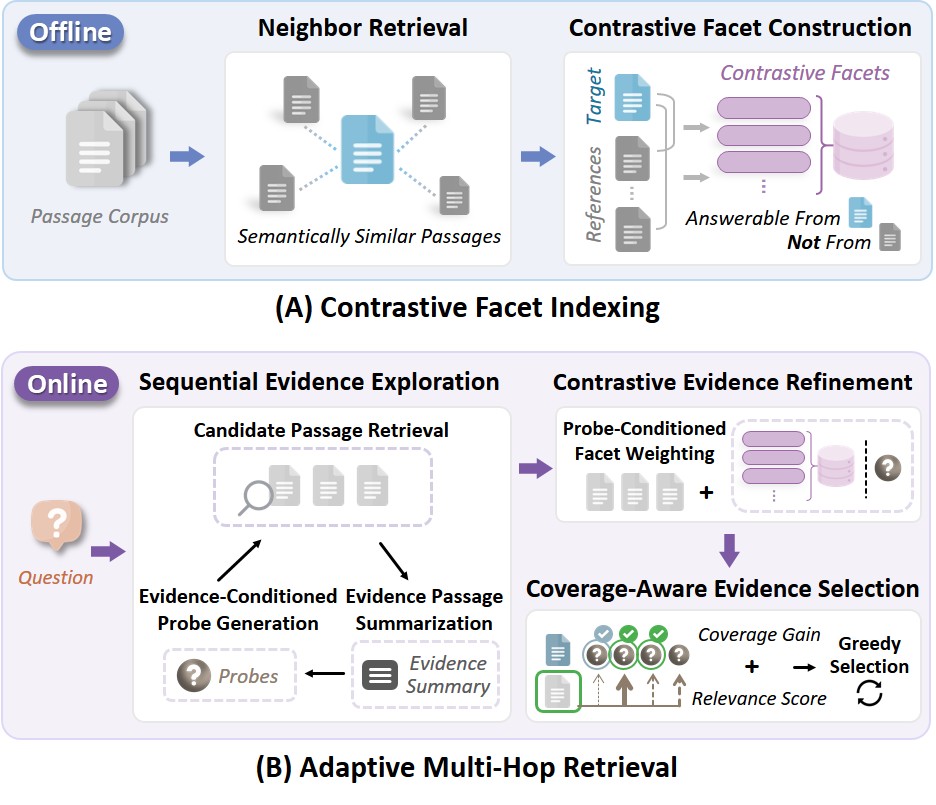}
  \caption{Overall pipeline of the proposed retrieval framework, consisting of (A) offline contrastive facet indexing and (B) online multi-hop retrieval.}
\end{figure}

\subsection{Pipeline Overview}

Our framework consists of two stages: offline contrastive facet indexing and online multi-hop retrieval. 
\begin{itemize}  
    \item \textit{During the offline stage}, we construct passage-specific contrastive facets as natural-language queries that distinguish each passage from its semantically similar neighbors. These facets are stored as a contrastive index and reused during inference without requiring additional training. 
    \item \textit{During the online stage}, the framework performs sequential evidence exploration, iteratively generating probes based on retrieved evidence to guide subsequent retrieval. After exploration, the accumulated candidates are refined using passage-specific contrastive facets and ranked with a coverage-aware strategy that favors complementary evidence across the generated probes.
\end{itemize}

\subsection{Proposed Framework}

\subsubsection{Contrastive Facet Construction \& Indexing}

For each passage $d_i$, we first retrieve a set of semantically similar neighboring passages, denoted by $R_i \subset \mathcal{D}\setminus\{d_i\}$. These neighboring passages serve as contrastive references for identifying evidence that is distinctive to the target passage. For each reference passage $d_r \in R_i$, we provide an LLM with the target passage $d_i$ and the reference passage $d_r$. We operationalize a contrastive facet as a natural-language query that satisfies the following condition: \textit{each query should be answerable from the target passage $d_i$, but not answerable from the reference passage $d_r$}. 

We denote the set of contrastive facet queries generated for the target-reference pair $(d_i, d_r)$ as $\mathcal{C}_{i,r}=\{c_{i,r}^{(1)},\ldots,c_{i,r}^{(M_{i,r})}\}$. The contrastive facet set associated with $d_i$ is then defined as $\mathcal{C}_i=\bigcup_{d_r\in R_i}\mathcal{C}_{i,r}$. Each facet in $\mathcal{C}_i$ captures an information need that distinguishes the target passage from a semantically similar neighbor, providing passage-specific retrieval signals. The facets are generated once during offline indexing and stored for reuse at inference time.

\subsubsection{Sequential Evidence Exploration}

Given a multi-hop question $q$, the online retrieval stage performs $T$ rounds of sequential evidence exploration. Rather than generating all retrieval probes from the original question at once, our framework progressively updates its search direction using evidence retrieved in the preceding rounds.

Let $\mathcal{P}_t(q)$ denote the set of probes used at round $t$, and $\mathcal{A}_t(q)$ the accumulated candidate pool after round $t$. The first round uses the original question together with an initial set of probes generated from the question. At each round $t$, each probe $p\in\mathcal{P}_t(q)$ retrieves its top-$K_{\mathrm{cand}}$ passages from $\mathcal{D}$ according to semantic similarity. The candidate pool is updated as:
\begin{equation}
\mathcal{A}_t(q)=\mathcal{A}_{t-1}(q)\cup
\bigcup_{p\in\mathcal{P}_t(q)}
\operatorname{Ret}_{K_{\mathrm{cand}}}(p,\mathcal{D}),
\end{equation}
where $\mathcal{A}_0(q)=\varnothing$.

After each round $t<T$, we construct a compact evidence summary $e_t$ from the retrieved passages, conditioned on the original question $q$ and the current probes. The LLM is instructed to generate $e_t$ as a compact representation of the retrieved evidence for the next round. This summarization helps reduce the influence of redundant or irrelevant passage content when generating the next retrieval direction.

Using the original question $q$ and the evidence summary $e_t$, we perform evidence-conditioned query expansion. An LLM-based probe generator produces the next probe set $\mathcal{P}_{t+1}(q)$ by formulating search queries that target information not yet resolved by the retrieved context. The probes are encouraged to incorporate salient entities or relations from the retrieved evidence while pursuing evidence needed to answer the original question $q$. This allows the retrieval process to progressively uncover bridge entities and intermediate evidence that may not be explicitly expressed in the original question.

After $T$ rounds, the final candidate pool is $\mathcal{A}(q)=\mathcal{A}_T(q)$, and the complete probe set is $\mathcal{P}(q)=\bigcup_{t=1}^{T}\mathcal{P}_t(q)$. These are subsequently used for contrastive refinement and final evidence selection.

\begin{table*}[t!]
\centering
\caption{Evaluation results on the MuSiQue, HotpotQA, and 2WikiMultihopQA datasets across different retrieval methods. R@5 denotes Recall@5, and FSR@10 denotes Full Support Recall@10. Recall measures the proportion of supporting passages retrieved; FSR measures the proportion of questions for which all supporting passages are retrieved; MAP measures mean average precision over the retrieval results; and nDCG evaluates ranking quality by assigning higher weights to relevant passages appearing earlier in the ranking.}
\label{tab:retrieval_results}
\renewcommand{\arraystretch}{1.1}
\resizebox{1.0\textwidth}{!}{
\begin{tabular}{l cccc cccc cccc}
\toprule
& \multicolumn{4}{c}{\textbf{\textit{MuSiQue}}} 
& \multicolumn{4}{c}{\textbf{\textit{HotpotQA}}} 
& \multicolumn{4}{c}{\textbf{\textit{2WikiMultihopQA}}} \\
\cmidrule(lr){2-5} \cmidrule(lr){6-9} \cmidrule(lr){10-13}
\textbf{Method} 
& \textbf{R@5} & \textbf{MAP@10} & \textbf{nDCG@10} & \textbf{FSR@10} 
& \textbf{R@5} & \textbf{MAP@10} & \textbf{nDCG@10} & \textbf{FSR@10} 
& \textbf{R@5} & \textbf{MAP@10} & \textbf{nDCG@10} & \textbf{FSR@10} \\
\midrule
\midrule

BM25 & 34.58 & 28.91 & 39.18 & 10.20 & 66.75 & 57.03 & 66.61 & 62.80 & 54.60 & 47.71 & 59.16 & 28.70 \\
e5-large-v2 & 52.19 & 46.03 & 57.58 & 27.10 & 88.80 & 78.73 & 84.75 & 89.20 & 72.15 & 69.26 & 78.12 & 51.00 \\
bge-reranker & 59.08 & 53.61 & \underline{64.66} & 32.60 & \underline{92.00} & \underline{83.96} & \underline{88.46} & 92.20 & 72.25 & 71.15 & 79.51 & 49.20 \\
\midrule
\multicolumn{13}{l}{\textit{(Query Expansion)}} \\

HyDE & 54.17 & 46.88 & 58.72 & 31.30 & 82.35 & 72.19 & 79.34 & 82.60 & 70.00 & 64.57 & 74.22 & 47.30 \\
query2doc & 45.72 & 37.99 & 48.88 & 26.90 & 82.20 & 72.68 & 79.35 & 81.80 & 67.23 & 61.67 & 71.11 & 43.40 \\
LameR & 54.30 & 46.91 & 57.15 & \underline{39.20} & 90.65 & 78.90 & 84.36 & 91.80 & 84.52 & 79.15 & 84.83 & 74.30 \\
\midrule
\multicolumn{13}{l}{\textit{(Iterative Retrieval)}} \\

IRCoT & 50.66 & 42.50 & 59.19 & 37.90 & 88.75 & 79.13 & 85.55 & \underline{93.30} & 71.85 & 73.33 & 83.56 & 74.20 \\
Self-Ask & \underline{60.49} & \underline{54.63} & 64.42 & 34.50 & 91.55 & 80.25 & 85.67 & 92.10 & \underline{87.05} & \underline{84.97} & \underline{89.28} & \underline{75.60} \\
\midrule
\textbf{ours} & \textbf{72.93} & \textbf{63.31} & \textbf{72.67} & \textbf{57.00} & \textbf{95.40} & \textbf{85.41} & \textbf{89.47} & \textbf{96.20} & \textbf{95.25} & \textbf{91.16} & \textbf{93.66} & \textbf{91.50} \\
\bottomrule
\end{tabular}
}
\end{table*}

\subsubsection{Contrastive Evidence Refinement}

After sequential exploration, each candidate passage $d\in\mathcal{A}(q)$ is assigned a base relevance score according to its strongest matching probe:
\begin{equation}
s_{\mathrm{base}}(d)=\max_{p\in\mathcal{P}(q)}\operatorname{sim}(p,d).
\end{equation}
\\
This score captures the direct semantic similarity between a candidate passage and the evidence-seeking probes. In multi-hop retrieval, however, a passage may be relevant because it provides a specific bridge entity or intermediate relation, even when it is not globally similar to the original question. We therefore refine candidate relevance using the passage-specific contrastive facets constructed during offline indexing.

For a candidate passage $d$ with a contrastive facet set $\mathcal{C}_d=\{c_d^{(1)},\ldots,c_d^{(M_d)}\}$, we assign each facet a probe-conditioned weight based on its semantic similarity to probe $p$:
\begin{equation}
w_{p,d}^{(m)}=
\frac{\exp\!\big(\operatorname{sim}(p,c_d^{(m)})\big)}
{\sum_{m'=1}^{M_d}\exp\!\big(\operatorname{sim}(p,c_d^{(m')})\big)}.
\end{equation}
\\
We then construct a weighted facet representation:
\begin{equation}
\mathbf{b}_{p,d}=\sum_{m=1}^{M_d}w_{p,d}^{(m)}\,{\phi}(c_d^{(m)}).
\end{equation}

The probe-conditioned refined representation of passage $d$ is defined as $\widetilde{\mathbf{h}}_{p,d}
=
\operatorname{norm}\!\left({\phi}(d)+\alpha\,\mathbf{b}_{p,d}\right)$, where $\alpha$ controls the strength of the facet-based update and $\operatorname{norm}(\cdot)$ denotes $\ell_2$ normalization. The corresponding refined probe-passage score is $g(p,d)={\phi}(p)^\top \widetilde{\mathbf{h}}_{p,d}$.
\\
The contrastive relevance of a candidate passage is the strongest refined score over all probes:
\begin{equation}
s_{\mathrm{con}}(d)=\max_{p\in\mathcal{P}(q)}g(p,d).
\end{equation}
\\
The overall relevance score interpolates the base and contrastive scores:
\begin{equation}
s(d)=\beta\,s_{\mathrm{base}}(d)+(1-\beta)\,s_{\mathrm{con}}(d),
\end{equation}
where $\beta\in[0,1]$ controls the contribution of the original dense retrieval signal. This interpolation preserves the underlying dense retriever signal while allowing passage-specific contrastive facets to sharpen the ranking of candidates associated with intermediate evidence.

\subsubsection{Coverage-Aware Final Evidence Selection}

The final stage selects the top-$K_{\mathrm{final}}$ passages from the accumulated candidate pool $\mathcal{A}(q)$. Ranking candidates solely by individual relevance may repeatedly select passages addressing the same evidence-seeking intent~\cite{lee-etal-2025-shifting}. This is particularly undesirable in multi-hop retrieval, where the selected passages should jointly support different steps of the reasoning chain. We therefore adopt a coverage-aware greedy ranking objective.

Let $S$ denote the set of selected passages. For each probe $p$, we define its current coverage under $S$ as $\gamma_S(p)=\max_{d'\in S}g(p,d')$, with $\gamma_{\varnothing}(p)=0$. The marginal coverage gain of a candidate passage $d$ is then defined as: 
\begin{equation}
\Delta_{\mathrm{cov}}(d\mid S)=
\sum_{p\in\mathcal{P}(q)}
\max\!\big(0,\;g(p,d)-\gamma_S(p)\big).
\end{equation}
This term rewards candidates that provide additional support for probes not yet well covered by the selected set. At each step, we add the passage with the highest combined utility:
\begin{equation}
d^{*}=
\operatorname*{arg\,max}_{d\in\mathcal{A}(q)\setminus S}
\Big[
s(d)+\Delta_{\mathrm{cov}}(d\mid S)
\Big].
\label{eq8}
\end{equation}
This process is repeated until $|S|=K_{\mathrm{final}}$, and the order of selection determines the final ranking. By updating the coverage state after each selection, the greedy procedure progressively favors passages that provide complementary support for probes not yet well covered by the selected set. The resulting objective balances individual relevance with complementary coverage across the generated probes, encouraging the final evidence set to support diverse reasoning steps required for multi-hop question answering.

\section{Experiments}

\subsection{Experimental Settings}

We evaluate the proposed framework on three multi-hop QA datasets: MuSiQue~\cite{trivedi-etal-2022-musique}, HotpotQA~\cite{yang-etal-2018-hotpotqa}, and 2WikiMultihopQA~\cite{ho-etal-2020-constructing}, following prior work~\cite{luo2026gfm, trivedi2023interleaving, gutierrez2024hipporag}. We compare our approach against BM25~\cite{robertson2009probabilistic} as a sparse retrieval baseline, \texttt{e5-large-v2}\footnote{https://huggingface.co/intfloat/e5-large-v2}~\cite{wang2022text} as a dense retrieval baseline, and \texttt{bge-reranker-large}\footnote{https://huggingface.co/BAAI/bge-reranker-large}~\cite{10.1145/3626772.3657878} (hereafter referred to as \texttt{bge-reranker}) as a reranking baseline. We further include representative LLM-based query expansion methods, including HyDE~\cite{gao-etal-2023-precise}, Query2Doc~\cite{wang-etal-2023-query2doc}, and LameR~\cite{shen-etal-2024-retrieval}, as well as iterative retrieval methods, including IRCoT~\cite{trivedi2023interleaving} and Self-Ask~\cite{press-etal-2023-measuring}. 

Across all experiments, we use \texttt{e5-large-v2} as the text encoder and \texttt{Qwen2.5-7B}\footnote{https://huggingface.co/Qwen/Qwen2.5-7B-Instruct}~\cite{DBLP:journals/corr/abs-2412-15115} as the LLM. For a fair comparison, we reimplement all baselines under a unified experimental setting, using the same retrieval corpus and evaluation protocol, and the same encoder and LLM where applicable. For offline contrastive facet construction, we retrieve the top-10 nearest neighbors based on cosine similarity and use them as reference passages. For downstream QA evaluation, we use~\texttt{Llama-3.1-8B}\footnote{https://huggingface.co/meta-llama/Llama-3.1-8B-Instruct}~\cite{grattafiori2024llama} as the reader model and instruct it to answer each question using only the retrieved context provided to it. For our method, we set the number of retrieval rounds to $T=4$, with $K_{\mathrm{cand}}=10$ and $K_{\mathrm{final}}=10$. We set $\alpha=0.5$ and $\beta=0.5$. For the reranking baseline, we first retrieve 50 candidate passages using \texttt{e5-large-v2} and then rerank them to obtain the final top-$K_{\mathrm{final}}$ passages.

\subsection{Experimental Results}

\subsubsection{Main Results}

As shown in Table~\ref{tab:retrieval_results}, we report retrieval performance on MuSiQue, HotpotQA, and 2WikiMultihopQA. Overall, our method achieves the best performance across all evaluated datasets and metrics. The gains are observed in both recall-oriented metrics, such as R@5 and FSR@10, and ranking-oriented metrics, including MAP@10 and nDCG@10, indicating improvements in both evidence coverage and ranking quality.

The improvements are particularly pronounced in FSR@10, which measures whether the complete set of supporting passages is retrieved. On MuSiQue, our method achieves an FSR@10 of 57.00, substantially outperforming the best baseline score of 39.20. It also improves MAP@10 from 54.63 to 63.31 and nDCG@10 from 64.66 to 72.67, demonstrating that the framework not only retrieves more complete evidence but also ranks relevant passages more effectively. A similar trend is observed on 2WikiMultihopQA, where FSR@10 increases from 75.60 to 91.50, accompanied by consistent gains in R@5, MAP@10, and nDCG@10.

On HotpotQA, where several baselines already achieve relatively high retrieval performance, our method still obtains the highest score across all metrics, including an increase in FSR@10 from 93.30 to 96.20. Although the gains are smaller than those on MuSiQue and 2WikiMultihopQA, their consistency indicates that the proposed framework remains effective across datasets with varying retrieval characteristics.

\begin{table}[t]
\centering
\caption{Ablation study of the proposed components on MuSiQue.}
\label{tab:ablation_results}
\renewcommand{\arraystretch}{1.1}
\resizebox{1.0\columnwidth}{!}{
\begin{tabular}{l ccc}
\toprule
\textbf{} & \textbf{R@10} & \textbf{nDCG@10} & \textbf{FSR@10} \\
\midrule \midrule
e5-large-v2 & 59.93  & 57.58 & 27.10 \\

\textbf{ours} & \textbf{80.41} & \textbf{72.67} & \textbf{57.00} \\
\midrule
\multicolumn{4}{l}{\textit{(Ablation Variants)}} \\
\quad w/o Coverage-aware Final Ranking & 78.83 & 71.12 & 54.60 \\
\quad w/o Contrastive Evidence Refinement & 79.72 & 72.38 & 55.60 \\
\quad w/o Sequential Evidence Exploration & 68.75 & 65.98 & 36.50 \\
\bottomrule
\end{tabular}
}
\end{table}

\subsubsection{Ablation Study}

As presented in Table~\ref{tab:ablation_results}, we report the ablation results on MuSiQue to assess the contribution of each major component. In \textit{w/o Sequential Evidence Exploration}, we retain only the initial round retrieval while removing subsequent evidence-conditioned retrieval rounds. In \textit{w/o Coverage-aware Final Ranking}, the accumulated candidates are ranked solely by the overall relevance score, without the coverage-aware selection. In \textit{w/o Contrastive Evidence Refinement}, the passage-specific facet-based refinement is removed, and candidate relevance is based on the base probe-passage relevance score. Removing any component reduces R@10, nDCG@10, and FSR@10 relative to the full model.

Sequential evidence exploration has the largest effect across all three metrics. Removing it decreases R@10 from 80.41 to 68.75, nDCG@10 from 72.67 to 65.98, and FSR@10 from 57.00 to 36.50. The largest decrease occurs in FSR@10, indicating that iterative exploration through subsequent evidence-conditioned retrieval rounds is particularly important for recovering the complete set of supporting passages. The substantial decreases in R@10 and nDCG@10 further show that sequential exploration also contributes to overall supporting-passage recall and ranking quality.

Coverage-aware final ranking provides smaller but consistent gains, especially on FSR@10 and nDCG@10. Removing it decreases FSR@10 from 57.00 to 54.60, while R@10 and nDCG@10 decrease from 80.41 to 78.83 and from 72.67 to 71.12, respectively. This pattern is consistent with the intended role of coverage-aware selection in improving the composition of the final evidence set so that the required supporting passages are jointly represented. Contrastive evidence refinement yields a modest improvement: removing it reduces R@10 from 80.41 to 79.72, nDCG@10 from 72.67 to 72.38, and FSR@10 from 57.00 to 55.60. This indicates that contrastive refinement provides a complementary relevance signal for distinguishing candidate passages after sequential exploration, thereby improving complete-support retrieval.

\begin{table}[t!]
\centering
\caption{Downstream QA performance using the top-5 retrieved passages.}
\label{tab:downstream_qa_top5}
\renewcommand{\arraystretch}{1.0}
\resizebox{1.0\columnwidth}{!}{
\begin{tabular}{l cc cc cc}
\toprule
& \multicolumn{2}{c}{\textbf{\textit{MuSiQue}}}
& \multicolumn{2}{c}{\textbf{\textit{HotpotQA}}}
& \multicolumn{2}{c}{\textbf{\textit{2WikiMultihopQA}}} \\
\cmidrule(lr){2-3} \cmidrule(lr){4-5} \cmidrule(lr){6-7}
\textbf{} 
& \textbf{EM} & \textbf{F1}
& \textbf{EM} & \textbf{F1}
& \textbf{EM} & \textbf{F1} \\
\midrule \midrule
BM25 & 8.1 & 15.39 & 35.0 & 47.59 & 20.3 & 26.94 \\
e5-large-v2 & 14.1 & 23.61 & 46.7 & 62.09 & 32.0 & 39.01 \\
bge-reranker & 13.6 & 24.84 & 49.1 & 64.55 & 36.2 & 43.10 \\
\midrule
HyDE & 16.5 & 26.11 & 43.2 & 58.75 & 32.1 & 39.24 \\
query2doc & 14.9 & 21.14 & 42.0 & 57.48 & 31.7 & 38.90 \\
LameR & \underline{20.5} & \underline{31.67} & \underline{49.8} & \underline{66.45} & 40.2 & 48.59 \\
\midrule
IRCoT & 13.5 & 22.20 & 47.1 & 62.80 & 31.3 & 38.43 \\
Self-Ask & 20.0 & 30.21 & 49.1 & 65.21 & \underline{42.2} & \underline{50.59} \\
\midrule
\textbf{ours} 
& \textbf{24.0} & \textbf{36.61}
& \textbf{51.8} & \textbf{68.80}
& \textbf{44.2} & \textbf{53.87} \\
\bottomrule
\end{tabular}
}
\end{table}

\subsubsection{Downstream QA Performance}

Table~\ref{tab:downstream_qa_top5} reports downstream QA performance when the reader model is provided with only the top-5 retrieved passages. Our method achieves the highest Exact Match (EM) and F1 scores across all three datasets among the compared methods, indicating that the improvements in retrieval quality are consistently reflected in downstream QA performance. 

On MuSiQue, our framework achieves 24.0 EM and 36.61 F1, improving over the best baseline scores of 20.5 EM and 31.67 F1. On 2WikiMultihopQA, EM increases from the best baseline score of 42.2 to 44.2, while F1 improves from 50.59 to 53.87.  On HotpotQA, where the baseline methods already achieve relatively strong QA performance, our method still improves EM from 49.8 to 51.8 and F1 from 66.45 to 68.80.

Overall, the gains are consistent across all three datasets, with the largest improvements observed on MuSiQue. These results suggest that retrieving more complete and better-ranked supporting evidence can improve downstream QA performance with a limited retrieval budget, although the resulting answer quality also depends on the reader model's ability to integrate and reason over the retrieved passages~\cite{zhang-etal-2026-failure}.

\section{Conclusion}

We present a training-free framework for multi-hop retrieval that integrates evidence-conditioned sequential exploration, passage-specific contrastive refinement, and coverage-aware final evidence selection. Rather than relying on a single retrieval intent or static one-shot query expansion, the framework progressively adapts its retrieval direction to the evidence retrieved at each step and selects complementary passages that jointly support the reasoning process. Experiments on three multi-hop QA datasets demonstrate consistent improvements in retrieval quality and downstream QA performance. Ablation results further show that sequential exploration plays the largest role in recovering complete supporting evidence, while contrastive refinement and coverage-aware ranking provide additional improvements. Overall, these findings suggest that conditioning retrieval on progressively discovered evidence and explicitly accounting for evidence coverage constitute an effective training-free approach to multi-hop retrieval.

Limitations include the computational cost of contrastive facet construction and sequential exploration. Facet construction requires LLM generation for each target–neighbor pair, despite limiting the number of neighbors in our experiments, while sequential exploration requires repeated LLM calls for evidence summarization and probe generation. These costs may increase offline processing and inference latency, particularly for large or frequently updated corpora.

\begin{acks}
This work was supported by the Institute of Information \& Communications Technology Planning \& Evaluation (IITP) grant funded by the Korea government (MSIT) [RS-2021-II211341, Artificial Intelligence Graduate School Program (Chung-Ang University)] and by the National Research Foundation of Korea (NRF) grant funded by the Korea government (MSIT) (RS-2025-00556246).
\end{acks}

\section*{GenAI Usage Disclosure}

For manuscript preparation, generative AI was used solely for grammatical correction and language editing. In the experiments, Qwen2.5-7B-Instruct was used within the proposed retrieval framework for contrastive facet query generation, evidence-seeking probe generation, and evidence summary generation. Llama-3.1-8B-Instruct was used as the reader model for downstream QA evaluation.

%% The next two lines define the bibliography style to be used, and
%% the bibliography file.
\bibliographystyle{ACM-Reference-Format}
\bibliography{sample-base}

\end{document}